\documentclass[conference]{IEEEtran}
\IEEEoverridecommandlockouts
\usepackage{cite}
\usepackage{amsmath,amssymb,amsfonts}
\usepackage{algorithmic}
\usepackage{graphicx}
\usepackage{textcomp}
\usepackage{xcolor}
\usepackage{array}  
\usepackage{graphicx}
\usepackage{subcaption} 
\usepackage{hyperref}

\def\BibTeX{{\rm B\kern-.05em{\sc i\kern-.025em b}\kern-.08em
    T\kern-.1667em\lower.7ex\hbox{E}\kern-.125emX}}
\begin{document}

\title{NoteBar: An AI-Assisted Note-Taking System for Personal Knowledge Management\\
\thanks{NoteBar, Google Cloud for Startups}
}

\author{\IEEEauthorblockN{1\textsuperscript{st} Josh Wisoff}
\IEEEauthorblockA{
\textit{NoteBar Research}\\
Schenectady, NY, USA \\
josh@okny.io}
\and
\IEEEauthorblockN{2\textsuperscript{nd} Yao Tang}
\IEEEauthorblockA{
\textit{University of Rochester}\\
Rochester, NY, USA \\
ytang49@ur.rochester.edu}
\and
\IEEEauthorblockN{3\textsuperscript{rd} Zhengyu Fang}
\IEEEauthorblockA{
\textit{Case Western Reserve University}\\
Cleveland, OH, USA \\
zxf177@case.edu}
\and
\IEEEauthorblockN{4\textsuperscript{th} Jordan Guzman}
\IEEEauthorblockA{
\textit{Skidmore College}\\
Saratoga Springs, NY, USA \\
jguzmanps4@gmail.com}
\and
\IEEEauthorblockN{5\textsuperscript{th} YuTang Wang}
\IEEEauthorblockA{\textit{University of Rochester} \\
Rochester, NY, USA \\
lulu104061@gmail.com}
\and
\IEEEauthorblockN{6\textsuperscript{th} Alex Yu}
\IEEEauthorblockA{
\textit{University of Rochester}\\
Rochester, NY, USA \\
alexchrisyu@gmail.com}
}

\maketitle

\begin{abstract}
Note-taking is a critical practice for capturing, organizing, and reflecting on information in both academic and professional settings. The recent success of large language models has accelerated the development of AI-assisted tools, yet existing solutions often struggle with efficiency. We present \textit{NoteBar}, an AI-assisted note-taking tool that leverages persona information and efficient language models to automatically organize notes into multiple categories and better support user workflows. To support research and evaluation in this space, we further introduce a novel persona-conditioned dataset of 3,173 notes and 8,494 annotated concepts across 16 MBTI personas, offering both diversity and semantic richness for downstream tasks. Finally, we demonstrate that NoteBar can be deployed in a practical and cost-effective manner, enabling interactive use without reliance on heavy infrastructure. Together, NoteBar and its accompanying dataset provide a scalable and extensible foundation for advancing AI-assisted personal knowledge management.
\end{abstract}


\section{Introduction}
Note-taking is a critical academic and professional practice, as efficient note-taking underpins attention, comprehension, and memory~\cite{piolat2005cognitive}. It functions both as an external record of information and as an active cognitive process that facilitates encoding and subsequent retrieval~\cite{piolat2005cognitive,mosleh2015challenges,jansen2017integrative}. With the increasing reliance on digital platforms, note-taking has shifted from traditional pen-and-paper to a broad range of digital systems~\cite{ward2003tool,srinivasa2021notelink,fang2021notecostruct,palani2021conotate}, which offer advantages such as searchability, portability, and integration with other applications. Nevertheless, this transition also introduces critical limitations~\cite{mosleh2015challenges}. Digital note-taking is inherently complex, as the processes of comprehension, summarization, and synthesis are difficult to support within digital environments, and existing systems often fail to capture these functions effectively. Moreover, many tools remain inefficient, constraining users to rigid, linear input rather than the flexible and spatially organized structures afforded by handwriting. Collectively, these challenges of complexity and inefficiency underscore the persistent gap between the affordances of current digital tools and the demonstrated cognitive benefits of traditional note-taking, highlighting the need for more robust and pedagogically grounded solutions.


Early digital note-taking systems such as StuPad~\cite{truong1999stupad}, NoteTaker~\cite{ward2003tool}, and Classroom Presenter~\cite{anderson2005use}, SmartNotes~\cite{banerjee2007segmenting}, Tsaap-Notes~\cite{silvestre2014tsaap}, facilitate note-taking but often offer fragmented functions, and provide only partial support for comprehension and synthesis~\cite{mosleh2015challenges}. 
More recently, AI-assisted methods have emerged: MeetingVis~\cite{shi2018meetingvis} focuses on real-time visualized summarization, NoTeeline~\cite {huq2025noteeline} expands user micronotes into full notes using LLMs, while GazeNoter~\cite{tsai2025gazenoter} combines augmented reality and gaze-based selection to co-pilot note generation. Therefore, AI-assisted note-taking represents a natural and promising direction for advancing the effectiveness of digital note-taking systems.


To address the limitations of existing note-taking systems, we propose \textit{NoteBar} (\url{https://notebar.ai/}), an AI-assisted platform that unifies classification, retrieval, and user interaction into a single workflow. Raw notes are transformed into structured representations for downstream classification, while retrieval supports contextual suggestions. A feedback mechanism is incorporated to enable iterative refinement and stronger alignment with user intent.
NoteBar combines persona-conditioned classification with retrieval-augmented suggestions and lightweight user-in-the-loop validation, enabling efficient organization of notes into actionable artifacts such as tasks, calendars, and playbooks. In doing so, NoteBar bridges the gap between the cognitive benefits of traditional note-taking and the scalability of AI copilots.

\noindent In summary, this work makes the following contributions:
\begin{itemize}
    \item We introduce \textit{NoteBar}, a novel AI-assisted note-taking system that unifies classification, retrieval, and user interaction within a single pipeline. 
    \item We design and release a persona-conditioned dataset of 3,173 synthetic notes with 8,494 concept annotations, enabling reproducible evaluation of multi-label note classification.
    \item We provide a deployment-ready architecture and a preliminary user study, showing that NoteBar improves efficiency and engagement in personal knowledge management workflows.
\end{itemize}

\section{Related Work}

\subsection{Note-taking Systems and Tools}

Traditional note-taking has long been dominated by pen-and-paper methods, valued for their flexibility, immediacy, and strong support for cognitive processes such as comprehension and memory~\cite{piolat2005cognitive}. With the increasing reliance on digital platforms, a wide range of digital tools such as Microsoft OneNote~\cite{pittman2007handwriting} and DyKnow~\cite{berque2006evaluation} have emerged to improve portability, searchability, and collaboration. While these systems alleviate storage and retrieval challenges, they often constrain users to linear or rigid interfaces and fall short in supporting the deeper integrative and reflective functions of note-taking~\cite{mosleh2015challenges}.  

To overcome these shortcomings, researchers have explored more interactive systems. For example, \textit{NoteLook}~\cite{chiu1999notelook} integrates live video with note-taking to help students align notes with lecture content, while \textit{Classroom Presenter}~\cite{anderson2005use} supports tablet-based ink annotations in collaborative classroom environments. Other systems, such as \textit{InkSeine}~\cite{hinckley2007inkseine} and \textit{VideoSticker}~\cite{cao2022videosticker}, extend traditional digital ink by embedding search, multimedia links, and timeline anchors directly into notes, thereby reducing the mechanical burden of manual writing and enhancing contextual grounding. Although these tools improve interactivity, they still require users to compose most of the content themselves, creating substantial cognitive load during fast-paced activities such as lectures or meetings.  

More recently, AI-assisted approaches have gained increasing attention. Systems such as \textit{MeetingVis}~\cite{shi2018meetingvis}, \textit{NoTeeline}~\cite{huq2025noteeline}, and \textit{GazeNoter}~\cite{tsai2025gazenoter} demonstrate how large language models and multimodal interaction can expand short inputs into rich notes, reduce cognitive effort, and enhance factual accuracy. For instance, NoTeeline expands user-provided micronotes into complete notes consistent with the user’s style, cutting writing effort nearly in half while maintaining factual correctness~\cite{huq2025noteeline}. These advances highlight AI-assisted note-taking as a promising direction, showing strong potential to complement human cognition and alleviate the limitations of traditional digital systems.

\subsection{Personal Knowledge Management and Routing}
Personal Knowledge Management (PKM) tools structure heterogeneous user notes into organized entities, enabling retrieval and suggestion workflows. Automatic routing of notes into semantic kinds parallels intent classification and tagging in note-taking systems~\cite{grippa2024obsidian}. While prior approaches often employ LLM prompting pipelines for tagging, these methods, though accurate, introduce high latency and cost and raise privacy concerns for sensitive data. Encoder-only models offer a practical alternative, supporting efficient on-device or CPU-based deployment~\cite{sanh2019distilbert,jiao2019tinybert,he2021debertav3}.

\subsection{Multi-Label Note Classification}

Multi-label note classification assigns multiple semantic labels to a single note, reflecting its overlapping intents and diverse functions. This setting is common in tagging, recommendation, and intent routing tasks~\cite{zhang2013review,read2011classifier}. Classical approaches include binary relevance, classifier chains, and neural architectures designed to handle long-tailed label distributions. In practice, short and informal notes often contain multiple intertwined intents, which makes multi-label classification particularly suitable. To address this, encoder-only transformers such as BERT~\cite{devlin2019bert} and DeBERTa~\cite{he2020deberta} provide effective sentence-level representations and remain strong baselines. Moreover, DeBERTa-v3~\cite{he2021debertav3} introduces disentangled attention and improved pretraining signals that enhance classification quality without adding inference complexity. These properties make encoder-only transformers, and DeBERTa-v3 in particular, a pragmatic choice for efficient and reliable note classification.



\begin{figure}[t]
    \centering
    \includegraphics[width=\linewidth]{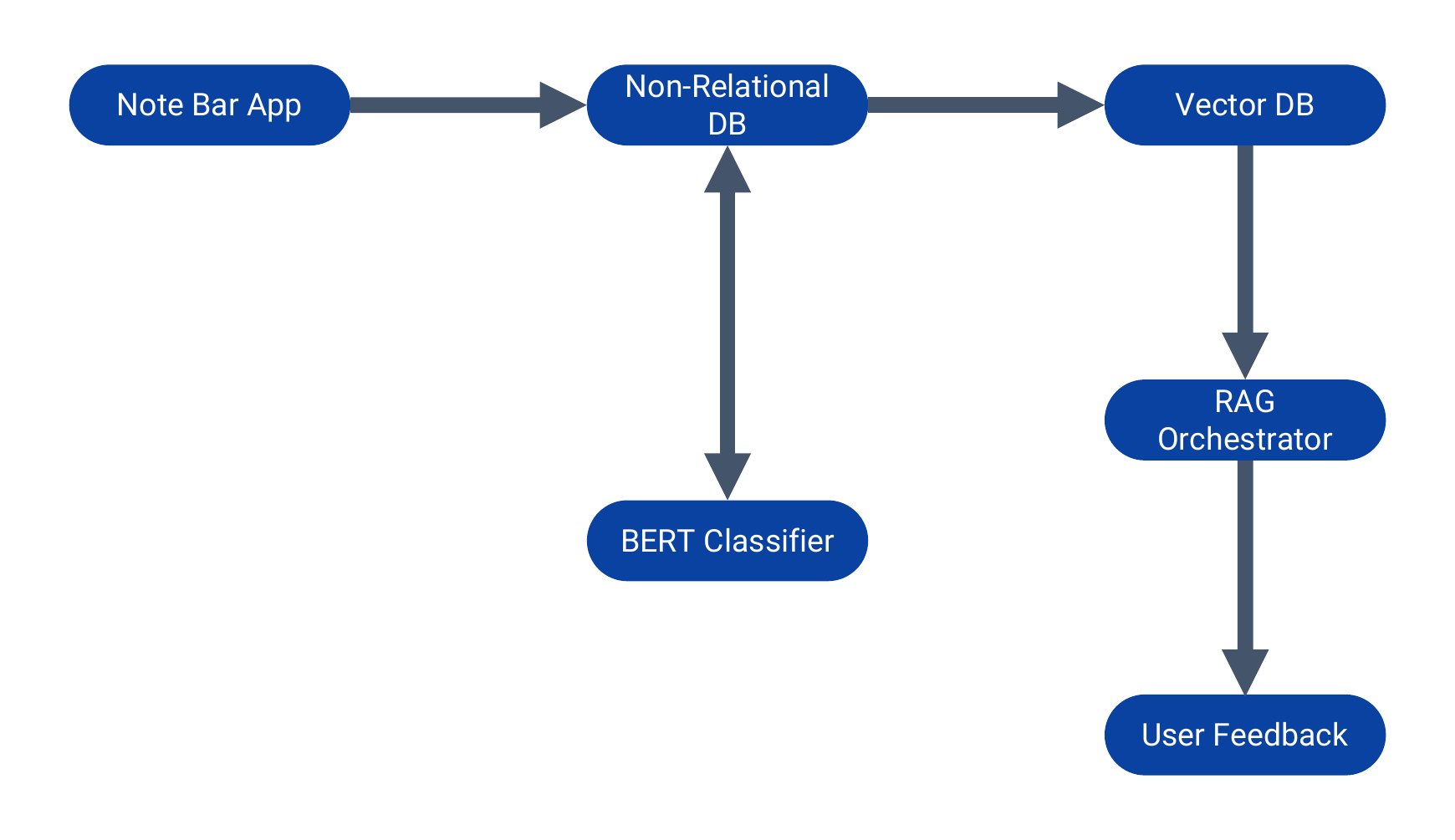}
    \caption{System architecture of \textit{NoteBar}. Notes created in the app are first stored in a non-relational database and classified by an encoder-only transformer model. Processed notes are embedded into a vector database, enabling retrieval through a RAG orchestrator. The orchestrator generates contextual suggestions, and user interactions provide feedback signals that refine both classification accuracy and retrieval quality.}
    \label{fig:notebar_system}
\end{figure}

\begin{figure*}[t]
    \centering
    \includegraphics[width=\textwidth]{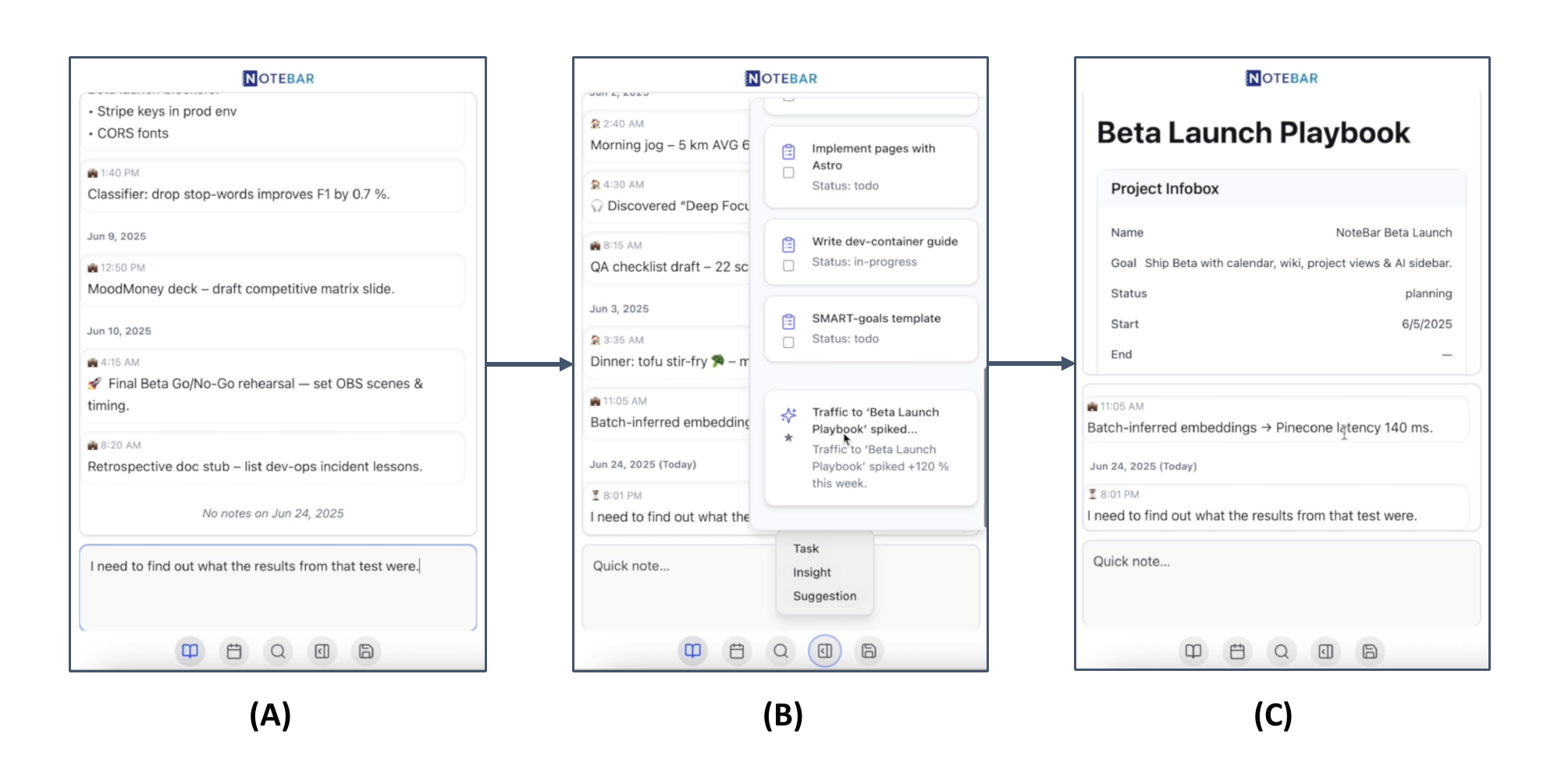}
     \caption{The NoteBar system pipeline. \textbf{(A)} Raw notes are first stored and vectorized. \textbf{(B)} Notes are routed by a BERT-based classifier with retrieval-augmented suggestions. \textbf{(C)} The system produces structured project artifacts (wikis), while user acceptance or edits are fed back to improve classification accuracy and retrieval quality over time.}
    \label{fig:notebar_pipeline}
\end{figure*}

\section{NoteBar}
\subsection{Design Principles}

The design of \textit{NoteBar} is guided by the need to bridge the cognitive benefits of traditional note-taking with the scalability and efficiency of AI-assisted systems. We articulate three core principles that shaped our system design.  

\textbf{Persona conditioning for disambiguation.} Effective note-taking support requires sensitivity to individual variation in writing style, intent, and role. Persona information, such as these stylistic and contextual cues, can be used to disambiguate utterances that would otherwise appear similar~\cite{ge2024scaling,jandaghi2023faithful}. Prior work has shown that conditioning generation or classification on persona tags helps capture systematic variation in language use. Building on this idea, NoteBar conditions classification on persona information to provide stable cues about how different users express concepts. This not only improves the separability of overlapping labels but also explains why: persona conditioning enables the system to normalize noisy or idiosyncratic phrasing into simpler, cleaner representations that highlight concept boundaries. As a result, semantically distinct concepts that may appear conflated in the raw note text can be more reliably routed to different categories, all while preserving the integrity of the underlying taxonomy.

\textbf{Multi-label routing with efficient transformers.} Notes often express multiple concepts simultaneously (e.g., \#task, \#insight, \#idea), making multi-label classification essential. To balance task quality with deployability, NoteBar formulates routing as a multi-label classification problem and employs encoder-only transformers (e.g., DeBERTa-v3~\cite{he2021debertav3}) as the backbone. This choice enables strong predictive performance while meeting strict latency and cost requirements through quantization and CPU-first serving.  

\textbf{User-in-the-loop feedback for robustness.} Automatic classification alone cannot fully capture user intent. To address this, NoteBar is designed to incorporate a feedback loop in which users can accept, reject, or edit system suggestions. While the current implementation does not yet realize this functionality, it represents an important component of our design principles and an active area of ongoing development.

\subsection{System Architecture}

The architecture of \textit{NoteBar} is designed as a modular pipeline that connects note capture, classification, retrieval, and eventually feedback, as illustrated in Figure~\ref{fig:notebar_system}. When a user creates a note within the application, the entry is stored in a non-relational database together with contextual metadata such as timestamp and device information (Figure~\ref{fig:notebar_pipeline}A). A BERT-based classifier is then applied to this stored content, performing multi-label routing that assigns each note to one or more semantic kinds. This classification step serves as the foundation for downstream workflows including task management, knowledge organization, and contextual retrieval.

\begin{figure}[t]
    \centering
    \includegraphics[width=0.85\linewidth]{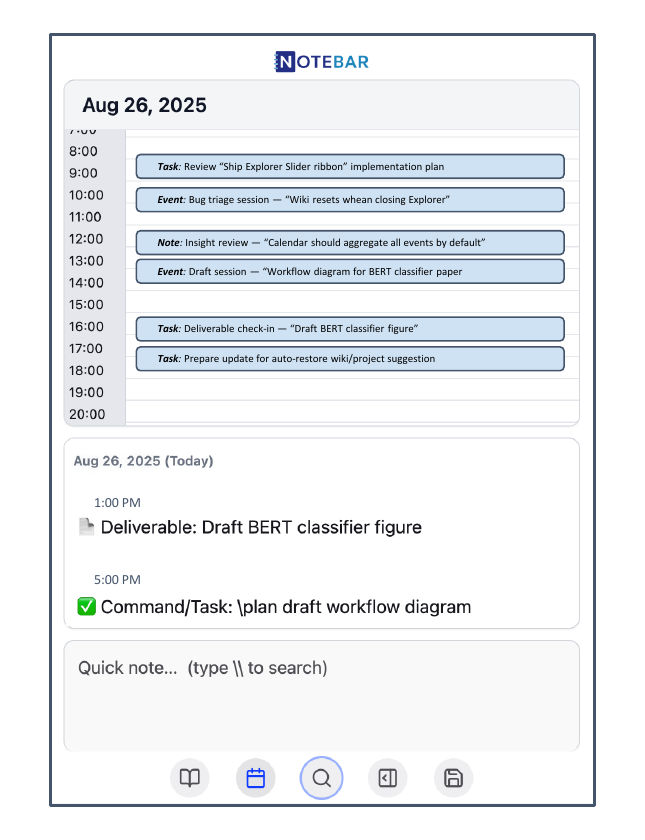}
     \caption{Calendar integration in \textit{NoteBar}. Notes are automatically transformed into tasks and scheduled calendar events. The day view organizes each entry into specific time slots (e.g., 9:00 AM task review, 10:00 AM bug triage, 1:00 PM draft session, 5:00 PM deliverable check-in), creating an actionable agenda directly derived from notebook entries.}
    \label{fig:notebar_calendar}
\end{figure}

\begin{figure}[t]
    \centering
    \includegraphics[width=0.85\linewidth]{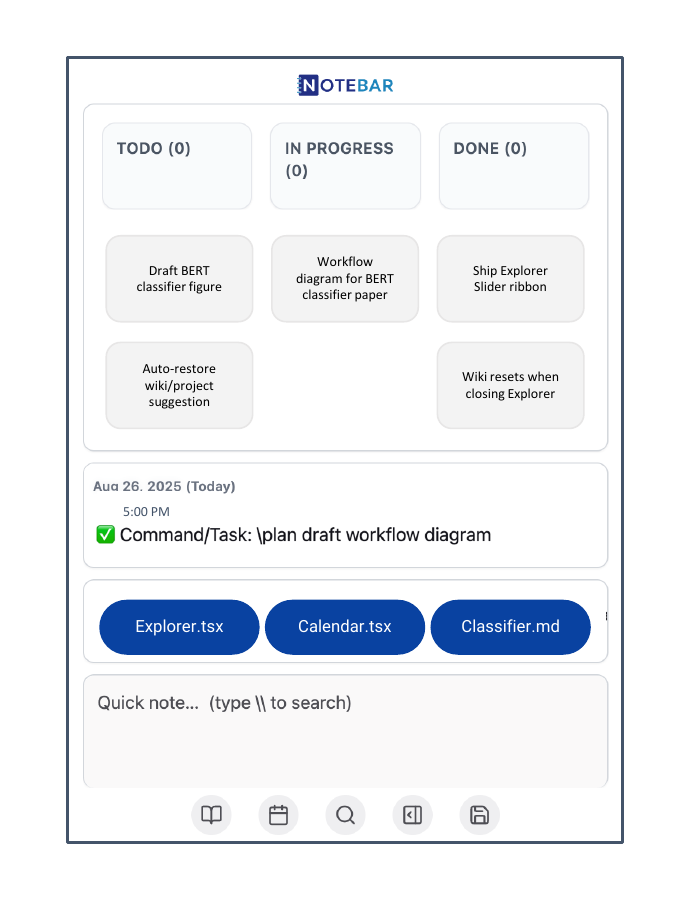}
     \caption{Kanban integration in \textit{NoteBar}. Notes are automatically transformed into project management tasks and organized by status. Items such as ``Review Explorer Slider ribbon'' and ``Bug triage — Wiki toggle'' appear under \textit{Done}, ``Draft workflow diagram'' remains in \textit{In Progress}, while ``Deliverable check-in — Draft BERT classifier figure'' and ``Prepare update for auto-restore wiki/project'' are listed as \textit{To Do}. This view illustrates how freeform notes are converted into structured artifacts that support task tracking within the RAG workflow.}
    \label{fig:notebar_kanban}
\end{figure}

To support retrieval-augmented suggestions, processed notes are further embedded and stored in a vector database. The vector database enables semantic similarity search, allowing the system to identify related notes, projects, or artifacts beyond exact keyword matching. A retrieval-augmented generation (RAG) orchestrator then leverages this index to assemble contextual information and produce candidate suggestions, such as linking a note to a calendar entry, playbook, or kanban task (Figure~\ref{fig:notebar_pipeline}B and Figure~\ref{fig:notebar_kanban}). In addition, actionable items such as deadlines or scheduled tasks can be automatically synchronized with the user’s calendar, ensuring that critical events are seamlessly integrated into their workflow (Figure~\ref{fig:notebar_calendar}).

Finally, NoteBar is designed to incorporate a user-in-the-loop feedback loop. In this envisioned component, suggestions could be accepted, edited, or dismissed by the user, and these interactions would be recorded as feedback signals (Figure~\ref{fig:notebar_pipeline}C). Such a mechanism would allow the system to adapt over time, refining both classification accuracy and retrieval quality while better aligning outputs with user intent. While still under development, this feedback loop represents an important direction for making NoteBar more adaptive and user-centered. Together, the modular components balance efficiency, scalability, and extensibility, supporting practical deployment in personal knowledge management.

\subsection{Implementation Details}

We implement \textit{NoteBar} as a lightweight yet extensible system that integrates note storage, transformer-based classification, and retrieval-augmented generation into a single pipeline. Notes are first stored in a non-relational database (Firebase Firestore) to provide flexible schema support for heterogeneous user inputs. Metadata such as timestamp, device, and project context are stored alongside raw text, enabling downstream modules to condition suggestions on contextual information.  

For classification, we adopt an encoder-only transformer architecture. A DeBERTa-v3-base~\cite{he2021debertav3} backbone is trained for multi-label routing. During training, we use AdamW with linear warm-up scheduling, a learning rate of $2\times 10^{-5}$, and batch size of 8, achieving stable convergence within 10 epochs.  

The vector database is implemented with Pinecone, storing embeddings of processed notes for efficient semantic retrieval. A RAG orchestrator is built on top of this index to assemble contextual suggestions. Suggestions are routed to the user interface, where they can be accepted, dismissed, or edited. While currently under design, these interactions are intended to be logged as feedback signals and later used to recalibrate classification thresholds, forming the basis of a lightweight user-in-the-loop adaptation mechanism.


\section{Experiments and Evaluation}
\subsection{Dataset}
Existing resources for evaluating note-taking systems and PKM workflows are limited. Most available corpora either focus on formal documents or lack the fine-grained annotations required for multi-label routing tasks. At the same time, collecting large-scale real user notes introduces privacy concerns and annotation challenges, making it difficult to study classification, retrieval, and suggestion in realistic PKM scenarios. To address this gap, we construct a synthetic dataset through an automated, agent-based pipeline designed to simulate realistic note-taking behaviors while ensuring both structural consistency and semantic fidelity (\url{https://github.com/Open-Knowledge-of-New-York-LLC/notebar-persona-dataset}).

\subsubsection{Data Generation Pipeline}
Our dataset is created using a four-stage pipeline which leverages \texttt{GPT-4o} as the underlying language model and integrates multiple agents to ensure contextual realism, structured annotations, and annotation reliability.

\paragraph*{Persona Agent (Note Generation).}
The pipeline begins with a persona-driven generation mechanism. Each persona is defined by a structured profile based on the 16 Myers--Briggs Type Indicator (MBTI) types, with contextual prompts that specify daily routines, interests, and writing styles. The Persona Agent generates notes that follow a fixed schema of \texttt{[Date]}, \texttt{[Time]}, \texttt{[Location]}, \texttt{[Device]}, \texttt{[Weather]}, and \texttt{[Note Content]}, where the first five fields mimic system-generated metadata and the final field reflects the persona’s individual voice. To simulate temporal continuity, each persona follows an eight-week plan that guides daily and weekly note generation. A RAG mechanism further incorporates relevant past entries to maintain historical coherence.
An illustrative example of a generated note is shown below:

\vspace{0.5em}
\begin{quote}
\raggedright
[2023-08-14][19:45][Navy Pier, Chicago][iPhone 15][Clear skies, 32$^\circ$C] \\
I took a long walk by the lake after work today. The sunset was calming, and it helped me reflect on my goals for the week. I realize I need to prioritize creative projects that bring me a sense of purpose, rather than getting lost in routine tasks.
\end{quote}
\vspace{0.5em}

This example, generated by an \texttt{INFP} persona, demonstrates how our design combines system-simulated contextual metadata with personality-driven note content, yielding entries that are both structured and realistic.

\paragraph*{Concept Router (Content Extraction).}
Generated notes are then processed by the Concept Router, which transforms free-form text into structured JSON annotations. The router extracts attributes such as note type flags, entities, and cognitive states, and assigns each note to a predefined taxonomy of \textit{kind} categories (e.g., task, insight, idea). Each concept is enriched with canonical analyses across five rhetorical dimensions (\textit{telos, logos, ethos, pathos, kairos}), providing annotations beyond surface-level labels.

\paragraph*{Annotation QA Agent (Validation and Refinement).}
To ensure annotation reliability, we design a two-stage quality assurance (QA) procedure that combines rule-based and LLM-based validation. In Stage I, schema, type, and range checks are applied to guarantee structural correctness. In Stage II, GPT-4o is used to verify canonical-score consistency and semantic plausibility; for critical issues, an automatic repair mechanism is triggered, while low-confidence cases are flagged for human inspection. This layered design enforces schema completeness, numerical validity, and semantic alignment with natural-language justifications, producing annotations that are both structurally consistent and semantically faithful. Table~\ref{tab:qa_checks} provides a concrete summary of these validation and refinement steps, showing how rule-based checks are complemented by LLM-based consistency checks and auto-fix mechanisms to balance structural rigor with semantic reliability.

\paragraph*{NoteBar Suggestions.}
Although optional for dataset construction, our pipeline also supports the generation of actionable suggestions. Notes processed through the concept router can be linked to downstream artifacts such as calendar events, playbooks, and kanban tasks. This functionality reflects how PKM systems operationalize raw notes into structured workflows.

\begin{table*}[]
\caption{Summary of validation and refinement steps in the Annotation QA Agent.}
\label{tab:qa_checks}
\begin{center}
\begin{tabular}{l|p{10cm}}
\hline
\textbf{Check Type} & \textbf{Action} \\
\hline
Schema validation (Rule-based) & Verify required keys are present; flag missing fields. \\
Type validation (Rule-based) & Ensure correct JSON datatypes (lists, floats); flag mismatches. \\
Range validation (Rule-based) & Confirm numeric scores lie in $[0,1]$; correct if out-of-range. \\
Canonical-score consistency (LLM-based) & Compare natural-language analysis vs. numeric scores; flag if discrepancy $\geq 0.25$. \\
Pre-label plausibility (LLM-based) & Check alignment of summary, entities, and scores with raw note. \\
Auto-fix policy & For major issues, invoke GPT-4o fixer; accept only valid JSON. \\
Edge case tagging & Flag low-confidence or failed checks for human inspection; attach QA insight. \\
\hline
\end{tabular}
\end{center}
\end{table*}

\subsection{Dataset Statistics}
The final dataset contains 3,173 notes and 8,494 concept annotations across 16 personas, of which 8,349 concepts passed QA validation. On average, each note is associated with 2.7 concepts. Persona-level distributions vary, with some types producing fewer than 60 notes and others exceeding 350, reflecting behavioral diversity and scenario-based prompts. 

\begin{figure}[t]
    \centering
    \includegraphics[width=\linewidth]{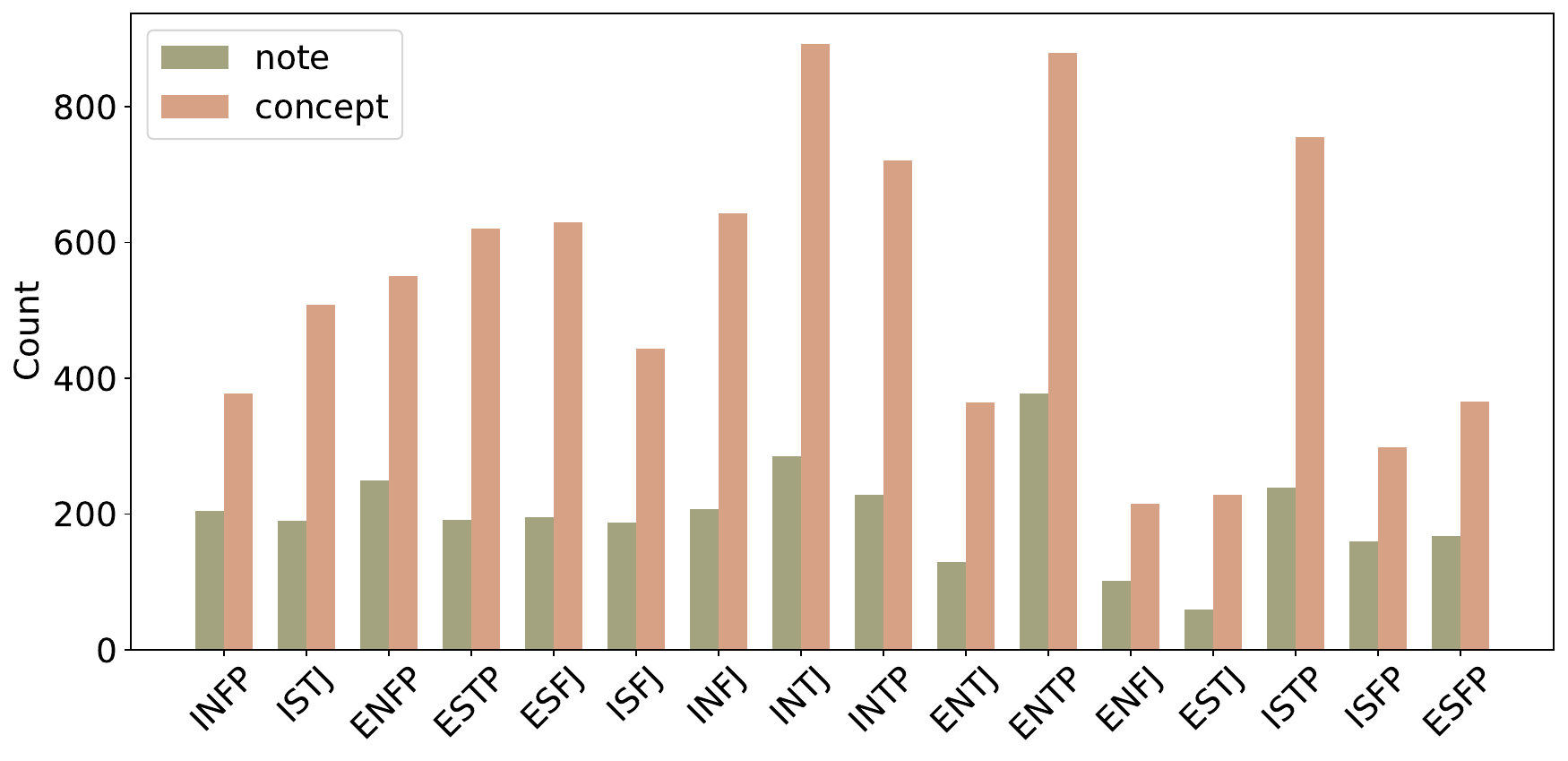}
    \caption{Distribution of notes and concepts across the 16 MBTI personas.}
    \label{fig:MBTI distribution}
\end{figure}

Table~\ref{tab:kind_dist} summarizes the distribution of semantic kinds. The dataset is dominated by \texttt{task} (5,170 instances), followed by \texttt{insight} (1,209) and \texttt{idea} (650), while rare categories such as \texttt{solution}, \texttt{ui\_action}, and \texttt{communication} appear only once. This long-tailed distribution mirrors real PKM settings, where frequent categories coexist with low-frequency but semantically important ones. Figure~\ref{fig:MBTI distribution} illustrates the distribution of notes and concepts across MBTI personas.

This dataset offers three distinctive features. First, persona conditioning ensures stylistic and intent diversity, reflecting realistic heterogeneity in note-taking practices. Second, the taxonomy exhibits a long-tailed structure, which presents challenges more representative of real PKM workloads than uniformly distributed corpora. Third, annotations combine surface-level metadata with rich semantic dimensions, enabling both classification and retrieval-based applications. Together, these characteristics make the dataset a valuable resource for evaluating models and systems aimed at bridging the gap between raw note-taking and actionable personal knowledge management.

\begin{table}[htbp]
\caption{Distribution of semantic kinds in the dataset.}
\label{tab:kind_dist}
\begin{center}
\begin{tabular}{l|r}
\hline
\textbf{Kind} & \textbf{Count} \\
\hline
task & 5170 \\
insight & 1209 \\
idea & 650 \\
suggestion & 394 \\
theme & 254 \\
goal & 202 \\
risk & 158 \\
requirement & 130 \\
decision & 51 \\
fact & 37 \\
tool\_feature & 35 \\
habit & 21 \\
draft & 16 \\
artifact & 7 \\
event & 5 \\
strategy & 4 \\
activity & 3 \\
solution & 1 \\
ui\_action & 1 \\
communication & 1 \\
\hline
\end{tabular}
\end{center}
\end{table}

\subsection{Experimental Setup}

We conduct experiments on the synthetic dataset introduced earlier, which contains 3,173 notes and 8,494 annotated concepts generated across 16 MBTI-based personas. Each note is associated with one or more semantic \textit{kind} labels (e.g., \texttt{task}, \texttt{insight}, \texttt{idea}), making the classification task inherently multi-label in nature. The distribution of labels is long-tailed, with frequent categories such as \texttt{task} coexisting with rare but semantically important ones (e.g., \texttt{solution}, \texttt{communication}). This imbalance presents additional challenges for classification models.

\subsubsection{Task Definition} 
The goal of our model is to predict the correct set of \textit{kind} labels for each input note. We formulate this as a multi-label classification problem, where the model outputs a probability distribution over all candidate labels, followed by a thresholding mechanism to determine the final predicted set.

\subsubsection{Baselines} 
We employ the Hugging Face \texttt{AutoModel} framework as the encoder backbone and evaluate three widely used pre-trained Transformer models: (i) \texttt{bert-base-uncased}~\cite{devlin2019bert}, used as the baseline; (ii) \texttt{roberta-base}~\cite{liu2019roberta}, which leverages a larger pre-training corpus and dynamic masking; and (iii) \texttt{deberta-v3-base}~\cite{he2020deberta}, which incorporates disentangled attention and enhanced pre-training efficiency. To ensure fairness, all models are equipped with the same attention pooling layer and classification head.

\subsubsection{Evaluation Metrics} 
We report both Accuracy and F1-score as our primary metrics. Accuracy reflects overall correctness, while F1-score better captures performance under class imbalance by considering both precision and recall. Given the long-tailed label distribution, F1-score provides a more informative measure of model performance.

\subsubsection{Training Details} 
Models are fine-tuned with the AdamW optimizer and linear learning rate decay. We experiment with batch sizes $\{8, 16, 32\}$, learning rates $\{2 \times 10^{-5}, 3 \times 10^{-5}, 5 \times 10^{-5}\}$, and training epochs from 2 to 15. Hyperparameters are selected based on validation F1 performance.


\begin{figure}[t]
    \centering
    \includegraphics[width=\linewidth]{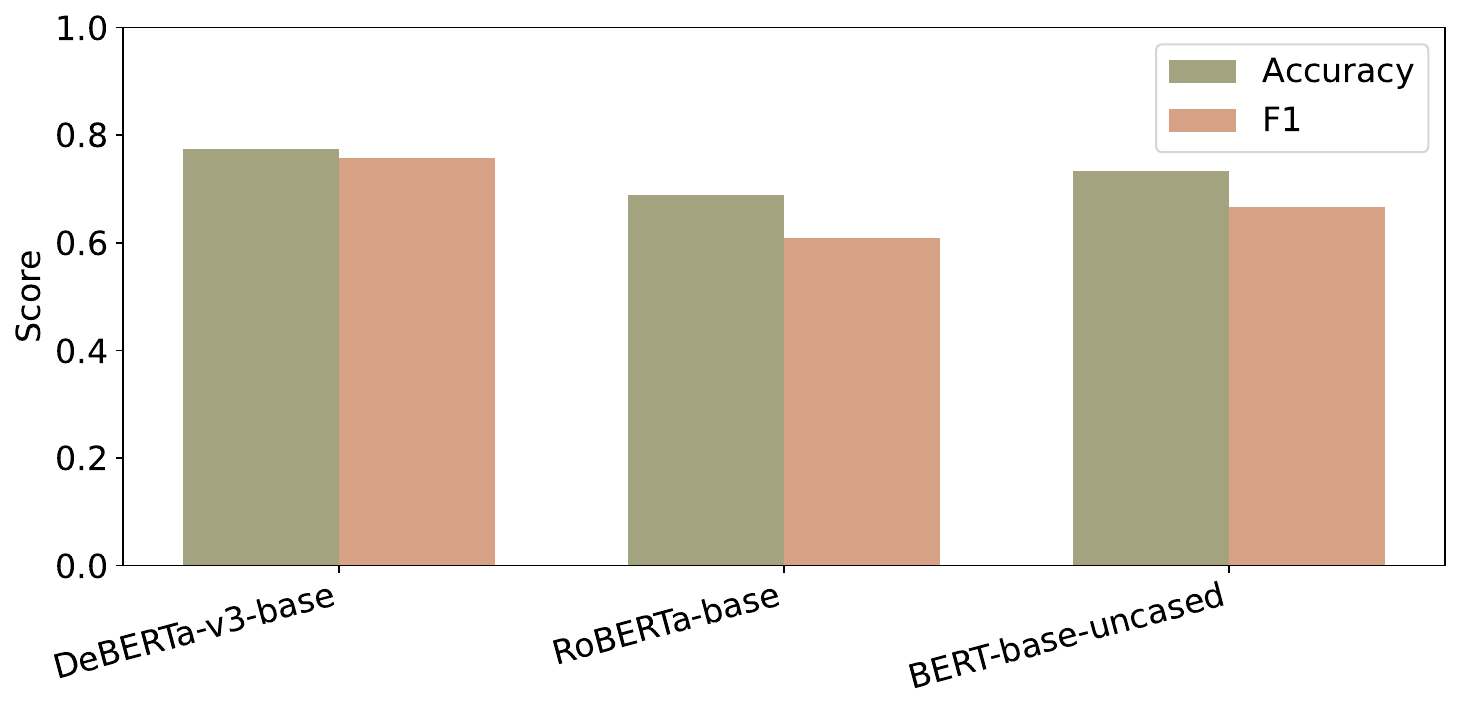}
    \caption{Performance Comparison of DeBERTa, RoBERTa, and BERT on Accuracy and F1.}
    \label{fig:overall_results}
\end{figure}



\begin{figure*}[t]
    \centering
    \begin{subfigure}{0.32\textwidth}
        \centering
        \includegraphics[scale=0.27]{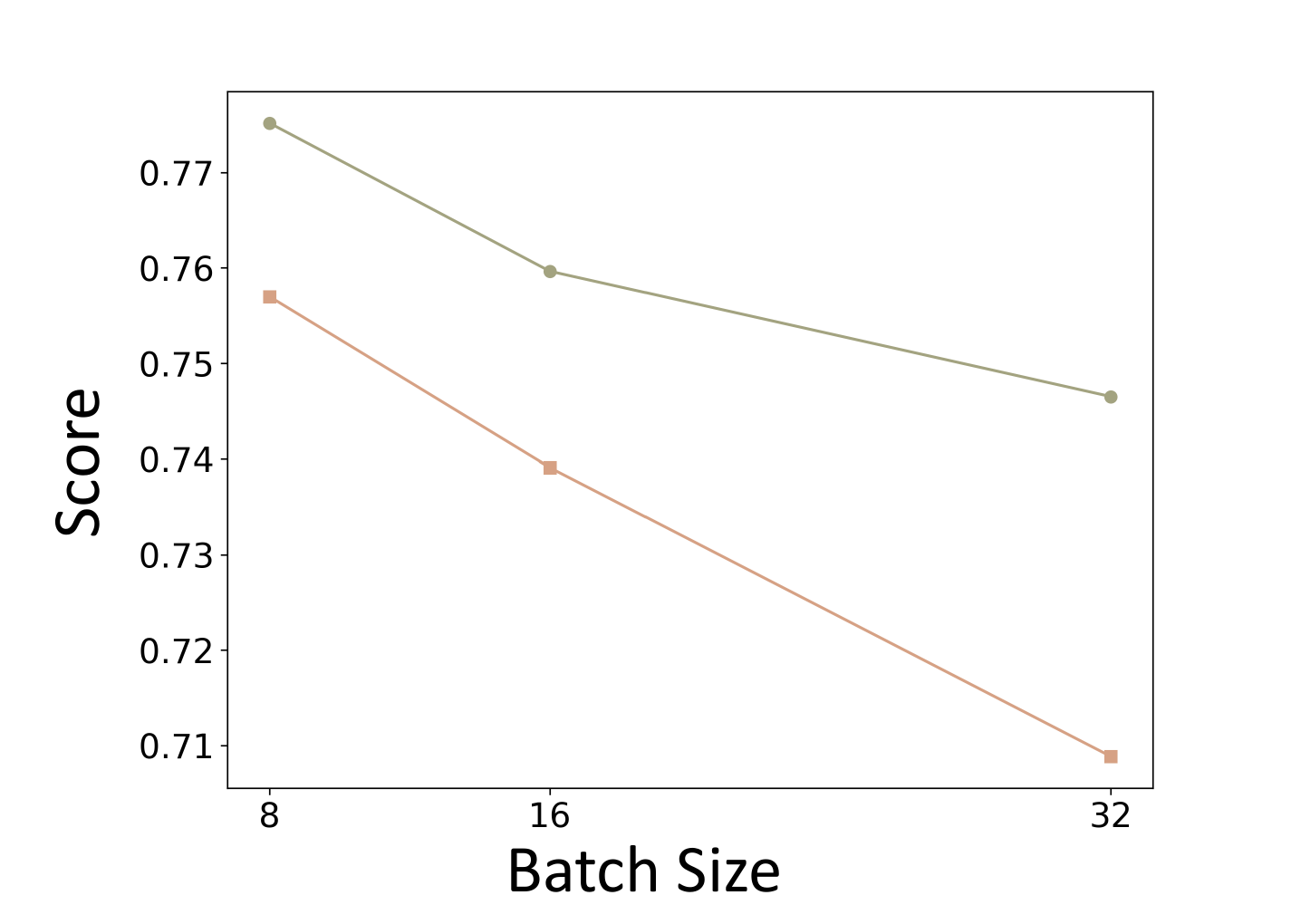}
        \caption{Effect of batch size.}
        \label{fig:bs_sensitivity}
    \end{subfigure}
    \hfill
    \begin{subfigure}{0.32\textwidth}
        \centering
        \includegraphics[scale=0.27]{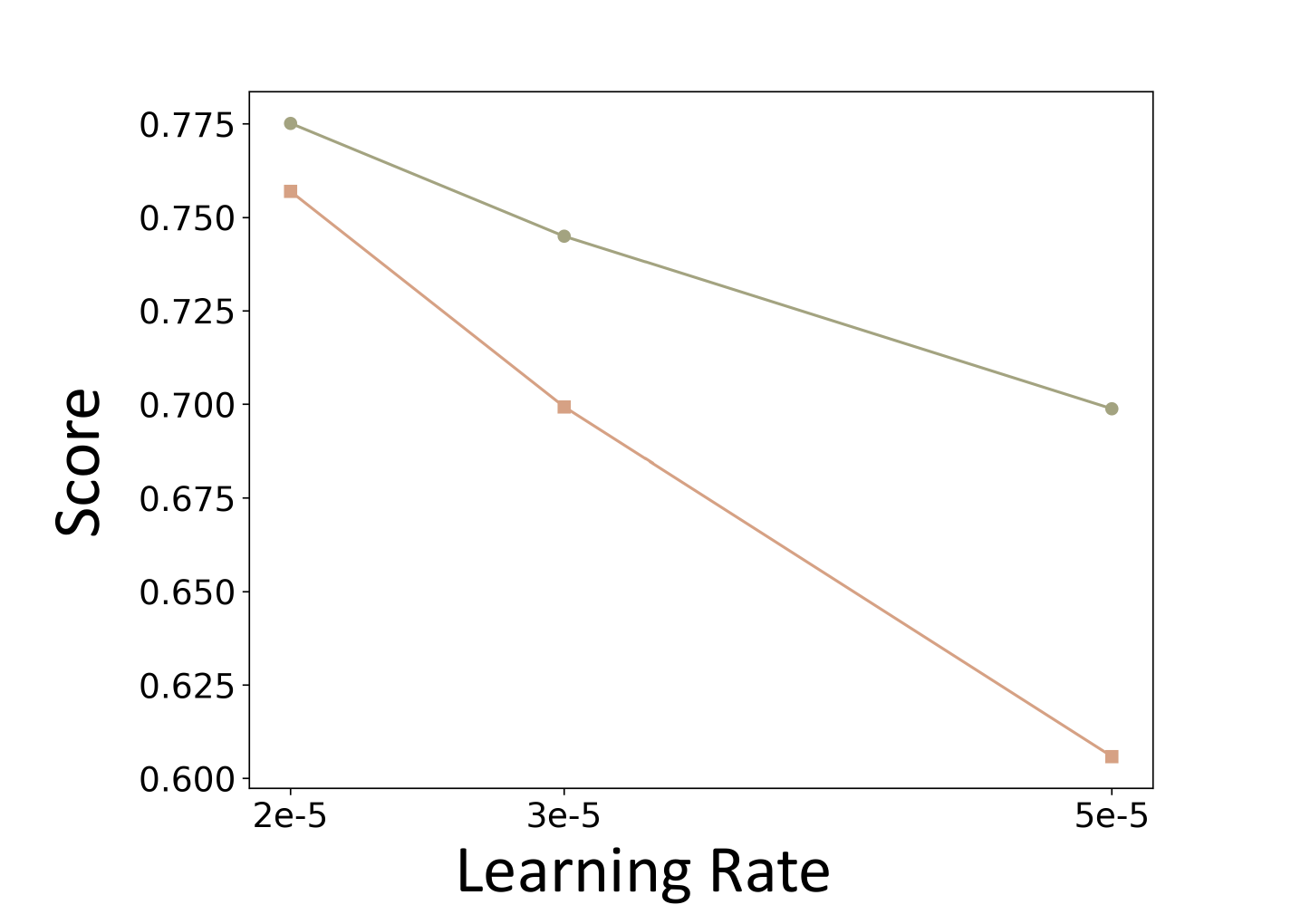}
        \caption{Effect of learning rate.}
        \label{fig:lr_sensitivity}
    \end{subfigure}
    \hfill
    \begin{subfigure}{0.32\textwidth}
        \centering
        \includegraphics[scale=0.27]{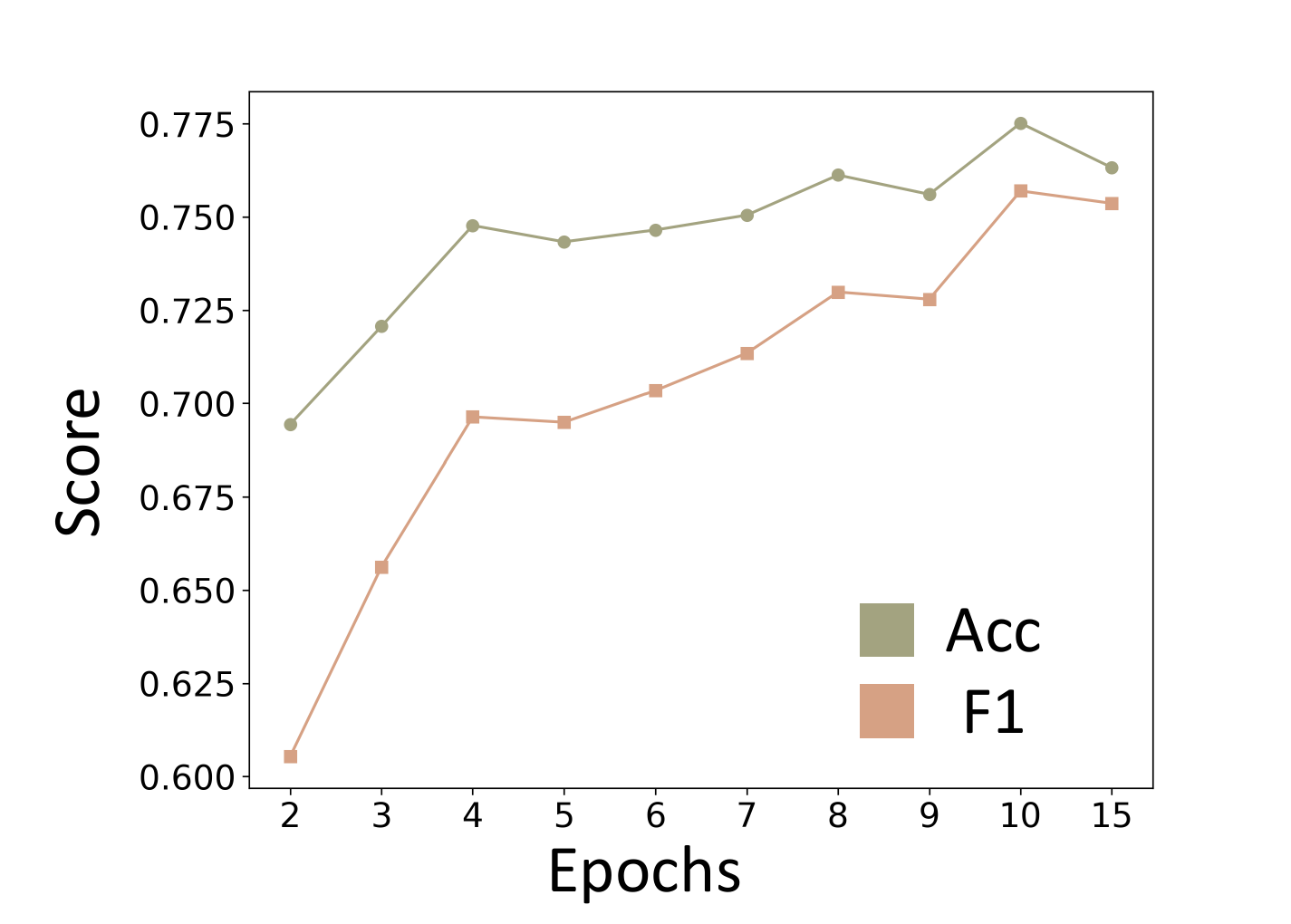}
        \caption{Effect of epochs.}
        \label{fig:epoch_sensitivity}
    \end{subfigure}
    \caption{Hyperparameter sensitivity analysis of DeBERTa-v3-base: effect of (a) batch size, (b) learning rate, and (c) number of epochs on accuracy and F1.}
    \label{fig:hyperparam_sensitivity}
\end{figure*}

\subsection{Results}

Figure~\ref{fig:overall_results} presents the best Accuracy and F1 achieved by each encoder under its optimal hyperparameters. Among the three backbones, \textit{DeBERTa-v3-base} clearly outperforms the alternatives, achieving an Accuracy of 0.78 and F1 of 0.76. \textit{BERT-base-uncased} performs moderately, while \textit{RoBERTa-base} yields the weakest results (Acc: 0.69, F1: 0.61). The performance gap highlights the effectiveness of disentangled attention and enhanced pretraining in DeBERTa for discriminating overlapping semantic kinds. By contrast, RoBERTa’s domain mismatch with our dataset appears to limit its effectiveness.

\subsubsection{Hyperparameter Sensitivity}
To better understand training dynamics, we perform a hyperparameter sensitivity analysis on \textit{DeBERTa-v3-base}, the best-performing model. We vary one hyperparameter at a time (batch size, learning rate, number of epochs) while fixing the others at their optimal values.

\paragraph{Batch Size}
As shown in Figure~\ref{fig:bs_sensitivity}, smaller batch sizes yield consistently stronger results. A batch size of 8 achieves the best performance (F1 = 0.757, Acc = 0.775), while larger sizes degrade performance. This suggests that frequent gradient updates help the model capture fine-grained variations in our long-tailed dataset.

\paragraph{Learning Rate}
Figure~\ref{fig:lr_sensitivity} demonstrates that a moderate learning rate of $2 \times 10^{-5}$ produces the highest Accuracy (0.775) and F1 (0.757). Increasing the rate to $3 \times 10^{-5}$ noticeably reduces performance, while $5 \times 10^{-5}$ destabilizes optimization, underscoring the need for conservative tuning in this task.

\paragraph{Number of Epochs}
As illustrated in Figure~\ref{fig:epoch_sensitivity}, performance improves steadily up to around 10 epochs, where the model reaches peak Accuracy (0.775) and F1 (0.757). Beyond this point, gains diminish and mild overfitting emerges, with slight drops in both metrics by epoch 15. This indicates that the model converges within 8–10 epochs.

\subsection{Summary of Findings}

\textbf{Obs. 1.} \textit{DeBERTa-v3-base} achieves the strongest performance across both Accuracy and F1, establishing it as the most effective backbone for our note classification task.

\textbf{Obs. 2.} Model stability and generalization are highly sensitive to hyperparameters. Small batch sizes and moderate learning rates consistently deliver superior convergence compared to larger batches or aggressive learning rates.

\textbf{Obs. 3.} Training exhibits a plateau after approximately 10 epochs. Beyond this point, additional training leads to diminishing returns or mild overfitting, highlighting a limited effective training window.

These observations emphasize both the inherent difficulty of our long-tailed multi-label dataset and the necessity of careful hyperparameter tuning to achieve reliable performance.

\section{Limitations and Future Plans}

\subsection{Limitations}
Our current system faces several limitations. First, the taxonomy of 20 semantic kinds, while covering common PKM workflows, is not exhaustive. Domain-specific contexts such as programming notes, research citations, or UI annotations remain outside the scope. The system also assumes English-language input, leaving multilingual note-taking unaddressed~\cite{conneau2019unsupervised}. Furthermore, the dataset exhibits a long-tailed distribution: frequent categories such as \texttt{task} dominate, whereas rare kinds (\texttt{solution}, \texttt{communication}) are severely underrepresented, resulting in high variance. Another limitation lies in data generation: although synthetic persona-driven notes enable broad coverage and controllability, they risk reinforcing templated patterns and may diverge from real-world user behavior. Finally, persona conditioning, while effective in improving label separability, may also introduce spurious correlations between persona styles and label distributions, raising fairness and generalization concerns.

\subsection{Future Plans}
To address these limitations, we plan to expand the taxonomy to incorporate domain-specific note types and extend evaluation to multilingual contexts. For the long-tailed label problem, we will explore active learning, targeted augmentation, and per-label calibration strategies to improve rare-label performance. To mitigate risks from synthetic data drift, we will gradually supplement training with opt-in, privacy-preserving real user notes. For fairness, we intend to conduct systematic audits (e.g., counterfactual persona swaps) and develop debiasing strategies that explicitly regularize against spurious persona–label correlations. Finally, we aim to complete the design of the user-in-the-loop feedback module, enabling accepted, rejected, or edited suggestions to be reintegrated into the training process for continual refinement. Together, these steps will broaden coverage, strengthen generalization, and ensure that NoteBar remains both practical and equitable in real-world PKM settings.

\section{Conclusion}
In this work, we presented \textit{NoteBar}, a persona-driven, encoder-only framework for multi-label concept routing in personal knowledge management. Using a 16-persona synthetic dataset (3,173 notes; 8,494 annotated concepts), we demonstrated that controlled data generation and QA pipelines can support reliable classification. Our experiments show that \textit{DeBERTa-v3-base} achieves strong performance on this challenging long-tailed task, while offering significant efficiency gains over GPT-based heuristics.

Beyond technical contributions, we introduced persona conditioning as a mechanism to improve separability among overlapping labels and showed how classification–retrieval integration supports practical PKM workflows. At the same time, several limitations remain: a restricted taxonomy, rare-label performance, potential drift in synthetic data, and fairness risks from persona conditioning. Moreover, while our architecture anticipates a user-in-the-loop feedback loop, this feature has not yet been implemented and will be a priority for future work. Moving forward, we plan to expand the taxonomy to domain-specific notes, extend to multilingual contexts, address long-tailed labels through active learning and augmentation, and incorporate privacy-preserving opt-in user data. Together, these directions aim to strengthen generalization and fairness, advancing toward scalable, privacy-preserving, and cognitively aligned note-taking support.

\section{Acknowledgments}
We thank the maintainers of BERT, the Hugging Face Transformers ecosystem, and LangChain. This research used cloud credits from Google Cloud for Startups. We acknowledge support from the NYS Center of Excellence — Goergen Institute for Data Science and AI, and from the Skidmore College Summer Experience Fund. We used the OpenAI API for text processing and generation during model development. All datasets used in this study were created by the authors.

\bibliographystyle{IEEEtran}
\bibliography{reference}

\begin{thebibliography}{10}
\providecommand{\url}[1]{#1}
\csname url@samestyle\endcsname
\providecommand{\newblock}{\relax}
\providecommand{\bibinfo}[2]{#2}
\providecommand{\BIBentrySTDinterwordspacing}{\spaceskip=0pt\relax}
\providecommand{\BIBentryALTinterwordstretchfactor}{4}
\providecommand{\BIBentryALTinterwordspacing}{\spaceskip=\fontdimen2\font plus
\BIBentryALTinterwordstretchfactor\fontdimen3\font minus \fontdimen4\font\relax}
\providecommand{\BIBforeignlanguage}[2]{{%
\expandafter\ifx\csname l@#1\endcsname\relax
\typeout{** WARNING: IEEEtran.bst: No hyphenation pattern has been}%
\typeout{** loaded for the language `#1'. Using the pattern for}%
\typeout{** the default language instead.}%
\else
\language=\csname l@#1\endcsname
\fi
#2}}
\providecommand{\BIBdecl}{\relax}
\BIBdecl

\bibitem{piolat2005cognitive}
A.~Piolat, T.~Olive, and R.~T. Kellogg, ``Cognitive effort during note taking,'' \emph{Applied cognitive psychology}, vol.~19, no.~3, pp. 291--312, 2005.

\bibitem{mosleh2015challenges}
M.~A. Mosleh, M.~S. Baba, S.~Malek, and M.~A. Alhussein, ``Challenges of digital note taking,'' in \emph{Advanced Computer and Communication Engineering Technology: Proceedings of ICOCOE 2015}.\hskip 1em plus 0.5em minus 0.4em\relax Springer, 2015, pp. 211--231.

\bibitem{jansen2017integrative}
R.~S. Jansen, D.~Lakens, and W.~A. IJsselsteijn, ``An integrative review of the cognitive costs and benefits of note-taking,'' \emph{Educational Research Review}, vol.~22, pp. 223--233, 2017.

\bibitem{ward2003tool}
N.~Ward and H.~Tatsukawa, ``A tool for taking class notes,'' \emph{International Journal of Human-Computer Studies}, vol.~59, no.~6, pp. 959--981, 2003.

\bibitem{srinivasa2021notelink}
R.~J. Srinivasa, S.~Dodson, K.~Seo, D.~Yoon, and S.~Fels, ``Notelink: A point-and-shoot linking interface between students' handwritten notebooks and instructional videos,'' in \emph{2021 ACM/IEEE Joint Conference on Digital Libraries (JCDL)}.\hskip 1em plus 0.5em minus 0.4em\relax IEEE, 2021, pp. 140--149.

\bibitem{fang2021notecostruct}
J.~Fang, Y.~Wang, C.-L. Yang, and H.-C. Wang, ``Notecostruct: Powering online learners with socially scaffolded note taking and sharing,'' in \emph{Extended Abstracts of the 2021 CHI Conference on Human Factors in Computing Systems}, 2021, pp. 1--5.

\bibitem{palani2021conotate}
S.~Palani, Z.~Ding, A.~Nguyen, A.~Chuang, S.~MacNeil, and S.~P. Dow, ``Conotate: Suggesting queries based on notes promotes knowledge discovery,'' in \emph{Proceedings of the 2021 CHI Conference on Human Factors in Computing Systems}, 2021, pp. 1--14.

\bibitem{truong1999stupad}
K.~N. Truong and G.~D. Abowd, ``Stupad: Integrating student notes with class lectures,'' in \emph{CHI'99 extended abstracts on Human factors in computing systems}, 1999, pp. 208--209.

\bibitem{anderson2005use}
R.~Anderson, R.~Anderson, L.~McDowell, and B.~Simon, ``Use of classroom presenter in engineering courses,'' in \emph{Proceedings Frontiers in Education 35th Annual Conference}.\hskip 1em plus 0.5em minus 0.4em\relax IEEE, 2005, pp. T2G--13.

\bibitem{banerjee2007segmenting}
S.~Banerjee and A.~I. Rudnicky, ``Segmenting meetings into agenda items by extracting implicit supervision from human note-taking,'' in \emph{Proceedings of the 12th international conference on Intelligent user interfaces}, 2007, pp. 151--159.

\bibitem{silvestre2014tsaap}
F.~Silvestre, P.~Vidal, and J.~Broisin, ``Tsaap-notes--an open micro-blogging tool for collaborative notetaking during face-to-face lectures,'' in \emph{2014 IEEE 14th International Conference on Advanced Learning Technologies}.\hskip 1em plus 0.5em minus 0.4em\relax IEEE, 2014, pp. 39--43.

\bibitem{shi2018meetingvis}
Y.~Shi, C.~Bryan, S.~Bhamidipati, Y.~Zhao, Y.~Zhang, and K.-L. Ma, ``Meetingvis: Visual narratives to assist in recalling meeting context and content,'' \emph{IEEE Transactions on Visualization and Computer Graphics}, vol.~24, no.~6, pp. 1918--1929, 2018.

\bibitem{huq2025noteeline}
F.~Huq, A.~Samee, D.~C.-E. Lin, A.~X. Tang, and J.~P. Bigham, ``Noteeline: Supporting real-time, personalized notetaking with llm-enhanced micronotes,'' in \emph{Proceedings of the 30th International Conference on Intelligent User Interfaces}, 2025, pp. 1064--1081.

\bibitem{tsai2025gazenoter}
H.-R. Tsai, S.-K. Chiu, and B.~Wang, ``Gazenoter: Co-piloted ar note-taking via gaze selection of llm suggestions to match users' intentions,'' in \emph{Proceedings of the 2025 CHI Conference on Human Factors in Computing Systems}, 2025, pp. 1--22.

\bibitem{pittman2007handwriting}
J.~A. Pittman, ``Handwriting recognition: Tablet pc text input,'' \emph{Computer}, vol.~40, no.~9, pp. 49--54, 2007.

\bibitem{berque2006evaluation}
D.~Berque, ``An evaluation of a broad deployment of dyknow software to support note taking and interaction using pen-based computers,'' \emph{Journal of Computing Sciences in Colleges}, vol.~21, no.~6, pp. 204--216, 2006.

\bibitem{chiu1999notelook}
P.~Chiu, A.~Kapuskar, S.~Reitmeier, and L.~Wilcox, ``Notelook: Taking notes in meetings with digital video and ink,'' in \emph{Proceedings of the seventh ACM international conference on Multimedia (Part 1)}, 1999, pp. 149--158.

\bibitem{hinckley2007inkseine}
K.~Hinckley, S.~Zhao, R.~Sarin, P.~Baudisch, E.~Cutrell, M.~Shilman, and D.~Tan, ``Inkseine: In situ search for active note taking,'' in \emph{Proceedings of the SIGCHI conference on human factors in computing systems}, 2007, pp. 251--260.

\bibitem{cao2022videosticker}
Y.~Cao, H.~Subramonyam, and E.~Adar, ``Videosticker: A tool for active viewing and visual note-taking from videos,'' in \emph{Proceedings of the 27th International Conference on Intelligent User Interfaces}, 2022, pp. 672--690.

\bibitem{grippa2024obsidian}
\BIBentryALTinterwordspacing
L.~Grippa. (2024) Obsidian ai tagger: simplify tagging in obsidian. GitHub. [Online]. Available: \url{https://github.com/lucagrippa/obsidian-ai-tagger}
\BIBentrySTDinterwordspacing

\bibitem{sanh2019distilbert}
V.~Sanh, L.~Debut, J.~Chaumond, and T.~Wolf, ``Distilbert, a distilled version of bert: smaller, faster, cheaper and lighter,'' \emph{arXiv preprint arXiv:1910.01108}, 2019.

\bibitem{jiao2019tinybert}
X.~Jiao, Y.~Yin, L.~Shang, X.~Jiang, X.~Chen, L.~Li, F.~Wang, and Q.~Liu, ``Tinybert: Distilling bert for natural language understanding,'' \emph{arXiv preprint arXiv:1909.10351}, 2019.

\bibitem{he2021debertav3}
P.~He, J.~Gao, and W.~Chen, ``Debertav3: Improving deberta using electra-style pre-training with gradient-disentangled embedding sharing,'' \emph{arXiv preprint arXiv:2111.09543}, 2021.

\bibitem{zhang2013review}
M.-L. Zhang and Z.-H. Zhou, ``A review on multi-label learning algorithms,'' \emph{IEEE transactions on knowledge and data engineering}, vol.~26, no.~8, pp. 1819--1837, 2013.

\bibitem{read2011classifier}
J.~Read, B.~Pfahringer, G.~Holmes, and E.~Frank, ``Classifier chains for multi-label classification,'' \emph{Machine learning}, vol.~85, no.~3, pp. 333--359, 2011.

\bibitem{devlin2019bert}
J.~Devlin, M.-W. Chang, K.~Lee, and K.~Toutanova, ``Bert: Pre-training of deep bidirectional transformers for language understanding,'' in \emph{Proceedings of the 2019 conference of the North American chapter of the association for computational linguistics: human language technologies, volume 1 (long and short papers)}, 2019, pp. 4171--4186.

\bibitem{he2020deberta}
P.~He, X.~Liu, J.~Gao, and W.~Chen, ``Deberta: Decoding-enhanced bert with disentangled attention,'' \emph{arXiv preprint arXiv:2006.03654}, 2020.

\bibitem{ge2024scaling}
T.~Ge, X.~Chan, X.~Wang, D.~Yu, H.~Mi, and D.~Yu, ``Scaling synthetic data creation with 1,000,000,000 personas,'' \emph{arXiv preprint arXiv:2406.20094}, 2024.

\bibitem{jandaghi2023faithful}
P.~Jandaghi, X.~Sheng, X.~Bai, J.~Pujara, and H.~Sidahmed, ``Faithful persona-based conversational dataset generation with large language models,'' \emph{arXiv preprint arXiv:2312.10007}, 2023.

\bibitem{liu2019roberta}
Y.~Liu, M.~Ott, N.~Goyal, J.~Du, M.~Joshi, D.~Chen, O.~Levy, M.~Lewis, L.~Zettlemoyer, and V.~Stoyanov, ``Roberta: A robustly optimized bert pretraining approach,'' \emph{arXiv preprint arXiv:1907.11692}, 2019.

\bibitem{conneau2019unsupervised}
A.~Conneau, K.~Khandelwal, N.~Goyal, V.~Chaudhary, G.~Wenzek, F.~Guzm{\'a}n, E.~Grave, M.~Ott, L.~Zettlemoyer, and V.~Stoyanov, ``Unsupervised cross-lingual representation learning at scale,'' \emph{arXiv preprint arXiv:1911.02116}, 2019.

\end{thebibliography}

\end{document}